\documentclass[a4paper,fleqn]{cas-sc}

\usepackage[authoryear,longnamesfirst]{natbib}

\usepackage{graphicx}
\usepackage{amsmath,amssymb}
\usepackage{booktabs}
\usepackage{algorithm}
\usepackage[noend]{algpseudocode}
\usepackage{subcaption}
\usepackage{tikz}
\usetikzlibrary{arrows.meta, positioning, calc, fit}

\begin{document}
\let\WriteBookmarks\relax
\def\floatpagepagefraction{1}
\def\textpagefraction{.001}

% Short title
\shorttitle{Automated Borehole Core Analysis}

% Short author
\shortauthors{Imdad et al.}

% Main title of the paper
\title[mode = title]{
Automated Borehole Core Analysis with Report-Derived Weak Labels and Supervised Crack Segmentation}

%
%Automated Crack Segmentation and Geological Attribute Extraction from Digital Borehole Log Reports}

% Authors
\author[1,5]{Usama Imdad}[orcid=0009-0008-4847-9918]
\cormark[1]
\credit{Conceptualization, Methodology, Software, Writing - original draft}
\ead{usama.imdad@lums.edu.pk}

\author[2]{Ali Khan}
\ead{ali.khan@sydneywater.com.au}

\author[4]{Luke Lu}
\ead{luke.lu@aurecongroup.com}

\author[1,5]{Zubair Khalid}[orcid=0000-0001-7875-4687]
\ead{zubair.khalid@lums.edu.pk}

\author[3]{Arif Mahmood}[orcid=0000-0001-5986-9876]
\ead{arif.mahmood@itu.edu.pk}

\affiliation[1]{organization={Lahore University of Management Sciences},
            country={Pakistan}}
\affiliation[2]{organization={Sydney Water},
            country={Australia}}
\affiliation[3]{organization={Information Technology University},
            country={Pakistan}}
\affiliation[4]{organization={Aurecon},
            country={Australia}}
\affiliation[5]{organization={PI-Neuron},
            country={Australia}}

\cortext[1]{Corresponding author}

% Abstract
\begin{abstract}
Borehole archives commonly contain core tray photographs and corresponding digital log reports, but no native pixel-level crack annotations. We investigate two complementary approaches for extracting defect-spacing information from these archives. First, structured spacing categories recovered from the report text layer provide weak interval-level labels for classification. A DINO encoder trained on unlabeled core crops supplies domain-specific representations, and a manually verified subset is used to identify label inconsistencies. Second, we manually annotate 5,087 extracted core-row images and evaluate fully supervised crack-segmentation models. Our gated U-Net combines PiDiNet edge maps with Mask R-CNN masks through a learned spatial gating mechanism. This configuration achieves an F1 score of 0.860 and a crack-class IoU of 0.754, the highest result among the evaluated segmentation configurations. Deterministic post-processing converts predicted crack locations into defect-spacing categories. Separate rule-based branches estimate core-relative bedding angles and lithological color descriptors; their predictions agree with log-report references on 75.4\% and 84.7\% of 1,200 evaluated images, respectively. Because these references are extracted from existing reports, the reported values measure agreement with recorded geological observations rather than independent physical accuracy. The resulting framework combines report-derived weak supervision for spacing classification with fully supervised segmentation for image-based crack localization.

\end{abstract}

% % Highlights
% \begin{highlights}
% \item Defect spacing labels are read from log reports at no annotation cost.
% \item Gated fusion of edge and instance masks reaches 0.86 F1 on borehole cracks.
% \item Hierarchical inference improves spacing classification under noisy pseudo-labels.
% \item Bedding angles and lithological colors are recovered as complementary attributes.
% \end{highlights}

% Keywords

\maketitle

\section{Introduction}
\label{sec:intro_problem}
Rock mass classification systems (RMCS), including RQD, RMR, and Q, summarize rock mass condition in indices that guide tunnel support selection, slope design, and excavation planning. These indices depend on the geometry, condition, and orientation of cracks in the rock \citep{alejano2025rock}. We use \emph{cracks (discontinuities)} consistently throughout this paper. We also retain the operator's column heading, \emph{defect spacing}, for the distance between successive natural cracks along the core. Defect spacing is a dominant term in these indices. Because the indices directly influence support decisions and costs, the quality of the underlying spacing measurements constrains the quality of the resulting engineering decisions.

In the archive considered here, a geologist lays out the recovered core, distinguishes natural cracks from drilling-induced cracks, and visually assigns a defect-spacing category based on professional judgement rather than measuring every crack-to-crack interval. This procedure is time-consuming and depends on subjective judgements about individual cracks and spacing categories. Consequently, defect spacing measurements show substantial inter-observer variability and limited reproducibility \citep{alejano2025rock}. Two geologists can derive different indices from the same core, and completed log reports are rarely revisited.

\begin{figure}[pos=tbp]
  \centering
  \begin{minipage}[c]{0.60\columnwidth}
    \centering
    \includegraphics[width=\linewidth]{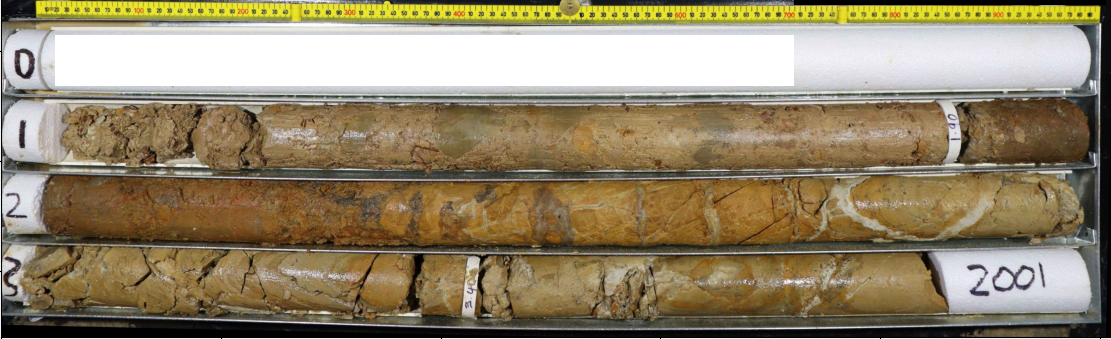}
  \end{minipage}\hfill
  \begin{minipage}[c]{0.38\columnwidth}
    \centering
    \includegraphics[width=\linewidth]{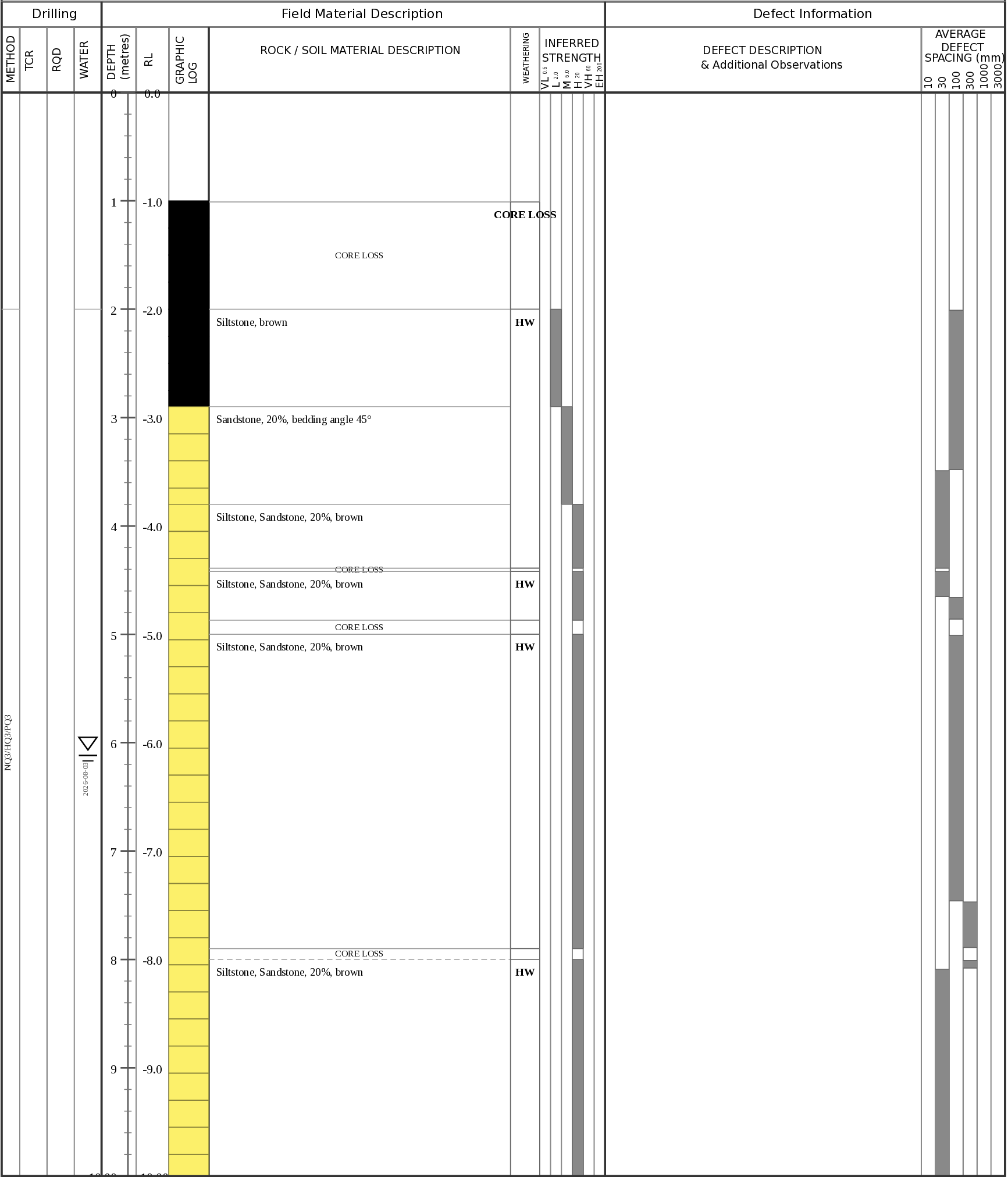}
  \end{minipage}
  \caption{The two input artifacts. Left: a core tray image containing the core images and the yellow ruler used to establish the pixel-to-millimetre scale. Right: the corresponding log report. Its ``AVERAGE DEFECT SPACING (mm)'' column contains the 10--3000\,mm categories recovered as pseudo-labels. The artifacts are matched by borehole identifier and depth interval. Solid blocks redact client-identifying information.}
  \label{fig:input_overview}
\end{figure}

Automation must operate on the two artifact types routinely retained for each
borehole. The first type comprises many core tray photographs
(Figure~\ref{fig:input_overview}, left). Each photograph contains three to five
approximately one-metre core rows, with a yellow ruler providing the
pixel-to-millimetre reference. The second type is a multi-page digital log
report (Figure~\ref{fig:input_overview}, right), produced by the operator's
logging software.

The core images present additional challenges. White depth labels are attached to the rock surface, core loss creates gaps, scratches can resemble cracks, and dark lithologies can reduce crack contrast almost entirely. Each extracted core image has an extreme aspect ratio of approximately $400\times5000$ pixels. The target cracks are therefore thin, elongated structures within a very wide image. Most importantly, \textbf{the archive contains no pixel-level ground truth}. The log report records only a defect spacing \emph{category} for each interval, so a conventional supervised formulation is not initially possible.

Existing work does not resolve this gap. CNN-based methods estimate RQD directly from core tray images \citep{alzubaidi2019rqd,alzubaidi2022rqd}. Related pipelines segment cracks and recover bedding angles from core tray images or unwrapped core images \citep{liu2025corefracture,alzubaidi2022unwrapped,zhou2023processes}. These methods assume purpose-captured images and, in most cases, pixel-level annotations. Crack segmentation is well established for concrete, pavement, and masonry \citep{hamishebahar2022review,liu2019deepcrack,kulkarni2022crackseg9k}, but these materials differ substantially from rock in texture and contrast. A method that converts this noisy, annotation-free archive into quantitative attributes is still missing. Section~\ref{sec:related} reviews the relevant literature.

We study two complementary formulations. The first treats the structured defect-spacing categories in the log reports as weak interval-level supervision for classification; it is intended for settings in which pixel-level annotations are unavailable. The second uses polygon masks created in this study to train fully supervised crack-segmentation models, after which deterministic geometric post-processing derives spacing from the predicted crack positions. The two formulations therefore address different annotation regimes and should not be interpreted as one weakly supervised segmentation method. Parallel rule-based branches estimate core-relative bedding angles and lithological color descriptors from the same extracted core images.

Our contributions are:
\begin{enumerate}
  \item We formulate defect-spacing analysis under two annotation regimes: report-derived weak supervision for spacing classification and fully supervised crack segmentation followed by geometric spacing computation.
  \item We evaluate a gated U-Net configuration that combines PiDiNet edge maps and instance-segmentation masks and achieves the highest crack-segmentation result among the evaluated configurations. We also document a setting in which fusion with a sparse CrackCLIP input degrades performance.
  \item We provide a deterministic post-processing procedure that converts ordered crack-mask components into physical spacing intervals using the image scale.
  \item We evaluate auxiliary rule-based procedures for core-relative bedding-angle estimation and lithological color description against values extracted from the corresponding log reports. 
\end{enumerate}

\section{Related Work}
\label{sec:related}
\textbf{Rock mass classification and automated core logging.} RMCS summarize rock quality for design and support decisions and remain central to geotechnical practice \citep{alejano2025rock}. These systems rely on crack spacing, crack condition, and structural orientation. This dependence motivates the automated extraction of crack geometry and defect spacing from core images. CNN-based methods estimate RQD directly from core tray images and reduce the subjectivity of manual logging \citep{alzubaidi2019rqd,alzubaidi2022rqd}.

\textbf{Core images and bedding angle estimation.} Core tray images and unwrapped core images support automated crack analysis with explicit geometric outputs. \citet{liu2025corefracture} compare direct crack segmentation with an indirect method that first identifies core blocks and then fits crack points to a plane. Workflows based on unwrapped core images and Mask R-CNN provide accurate crack detection and bedding angle estimates \citep{alzubaidi2022unwrapped}. Other studies estimate bedding angles from 3D digital core images through plane fitting \citep{zhou2023processes}. These studies motivate our emphasis on interpretable intermediate outputs.

\textbf{Digital rock and CT-based crack segmentation.} Digital rock physics and CT workflows demonstrate that segmentation quality directly affects crack quantification. Deep CNNs designed for rock crack images improve robustness to visual clutter \citep{byun2021deep}. Benchmarking studies also show that learning-based segmentation reduces user bias and improves downstream rock-physics estimates \citep{reinhardt2022drp}. Recent CT-based methods use deep networks to segment cracks and minerals in rock samples \citep{he2024resvgg} or combine tomography with neural segmentation to extract crack networks \citep{caputo2025sensors}. Engineering geology datasets further demonstrate that instance segmentation can automate crack identification at scale \citep{ji2025rockfracture}.

\textbf{Crack detection and segmentation in computer vision.} Crack segmentation has progressed from classical edge-based methods to learned dense prediction \citep{hamishebahar2022review}. DeepCrack introduced multi-scale feature aggregation for this task \citep{liu2019deepcrack}. Transfer learning and CNN-based segmentation have also been applied to masonry \citep{dais2021automatic}, while hierarchical feature aggregation improves pavement crack detection \citep{yang2019feature}. Contrastive self-supervision and mixed-label strategies address label scarcity \citep{song2024eswa,zhang2024automatic}, and segmentation remains an effective baseline for industrial surface images \citep{joshi2022automatic}. Architectures such as DepthCrackNet further improve segmentation accuracy \citep{saberironaghi2024depthcrack}. Vision-language methods, including CrackCLIP, extend weakly supervised segmentation through text prompts \citep{liang2025crackclip,radford2021learning}.

\textbf{Datasets, benchmarks, and learning with limited labels.} Data scarcity and heterogeneity remain important challenges. CrackSeg9k combines multiple datasets into a unified benchmark for cross-domain generalization \citep{kulkarni2022crackseg9k}. Crack500 and CrackForest provide established crack detection baselines \citep{zhang2016icip,shi2016rsf}, while CrackTree represents an earlier algorithmic approach \citep{zou2012cracktree}. Together, these datasets and methods inform the design of pipelines that tolerate label noise and domain shift.

\textbf{Self-supervision and geometric priors.} Self-supervised representation learning with DINO improves robustness when labels are sparse or noisy \citep{caron2021emerging}. Edge and line detectors such as PiDiNet and MLSD provide geometric priors for fine crack boundaries and core segmentation \citep{su2021pixel,gu2022towards}. These tools motivate hybrid pipelines that combine classical geometry with learned representations for interpretable core analysis.

\section{Data and Preprocessing}

\subsection{Study Area and Data Provenance}
\label{sec:provenance}
The corpus contains hundreds of boreholes gathered from all over Australia. Confidentiality restrictions prevent us
from identifying the operator and project. We have therefore removed all
client-identifying fields from the figures
(Figures~\ref{fig:input_overview} and~\ref{fig:extraction_pipeline}).

Each borehole is associated with one digital log report and many core tray
photographs. The reports store text and vector drawing commands, allowing the
recorded values to be read without optical character recognition. The tray
photographs are supplied as separate image files; although copies also appear
inside the reports, only the separately supplied photographs are used for image
analysis. Filenames encode the borehole identifier and depth interval, allowing
each photograph to be aligned with the corresponding report interval.

The cored intervals are dominated by interbedded siltstone and sandstone with
subordinate clay and span weathering grades from extremely to moderately
weathered. Core was recovered using an HQ3 triple-tube system with a nominal
61~mm core diameter. Logged depths range from approximately 10 to 60~m. Across
the corpus, multiple tray photographs per borehole yielded 5{,}087 retained
core-row images after the exclusions described in
Section~\ref{sec:seg_dataset}.

Following the operator's standard procedure, the logging geologist recorded
defect spacing, bedding angle, weathering, and lithological color in structured
log report columns. These include ``AVERAGE DEFECT SPACING (mm)'' with fixed
10--3000\,mm categories, crack descriptions with bedding angles, and material
descriptions with color. The recorded observations provide the reference for
every evaluation in Section~\ref{sec:results}. No independent measurements of
the core were available. Section~\ref{sec:limitations} discusses the
implications of this constraint.

\subsection{Core Image Extraction}

We extract core-row images in two stages. First,
Algorithm~\ref{alg:extract_area} detects and rectifies the tray region below the
yellow ruler. The tray image is converted to HSV color space and thresholded to
isolate the ruler. Morphological operations clean the mask, and aligned ruler
fragments are merged using vertical IoU and a convex hull. The fitted ruler
orientation is then used to correct image rotation. Second, filename metadata
provides the expected number of core rows. A line-segment detector
\citep{gu2022towards} identifies the horizontal row boundaries, and the expected
row count is used to select and order the resulting crops. All candidate crops
are manually inspected; non-core regions, placeholder rows, and extraction
failures are removed before downstream analysis.

\begin{algorithm}[htbp]
\caption{Tray Region Extraction and Rectification}
\label{alg:extract_area}
\begin{algorithmic}[1]
\Require RGB core tray image $I$
\Ensure Rectified image $I'$, tray region $R$, and ruler mask $M$
\State $I' \gets I$
\State Convert $I$ to HSV space, $I_{\text{HSV}} \gets \text{ConvertToHSV}(I)$
\State Define yellow color bounds, $L \gets [20, 100, 100]$ and $U \gets [30, 255, 255]$
\State Generate binary mask, $M \gets \text{InRange}(I_{\text{HSV}}, L, U)$
\State Apply morphological closing on $M$
\State Fill internal holes using contours
\State Extract all external contours $\mathcal{C}$ from $M$
\State Find contour $c_{\text{max}} \in \mathcal{C}$ with maximum width $w_{\text{max}}$
\If{$w_{\text{max}} < \frac{3}{4} \cdot \text{image width}$}
    \State Initialize set $\mathcal{S} \gets \{c_{\text{max}}\}$
    \For{each $c \in \mathcal{C}$}
        \State Compute vertical IoU between $c$ and $c_{\text{max}}$
        \If{IoU $> 0.3$}
            \State Add $c$ to $\mathcal{S}$
        \EndIf
    \EndFor
    \State Merge contours in $\mathcal{S}$ with a convex hull to obtain $c_{\text{merged}}$
    \State Set $c_{\text{max}} \gets c_{\text{merged}}$
\EndIf
\State Draw filled contour $c_{\text{max}}$ on a blank mask $R_{\text{mask}}$
\If{rotation correction is enabled}
    \State Fit rotated rectangle $r$ to $c_{\text{max}}$
    \State Compute rotation matrix $T$ from $r$
    \State Apply affine warp to $R_{\text{mask}}$ using $T$
    \State Update the region bounding box
\EndIf
\State Define $R$ as the rectangular tray region below $c_{\text{max}}$
\If{rotation correction is enabled}
    \State $I' \gets \operatorname{AffineWarp}(I,T)$
\EndIf
\State Crop $R$ from the rectified image $I'$
\State \Return $I'$, $R$, and $M$
\end{algorithmic}
\end{algorithm}

\begin{figure}[pos=tbp]
  \centering
  \includegraphics[width=0.48\columnwidth]{figs/redacted/input_image.png}\hfill
  \includegraphics[width=0.48\columnwidth]{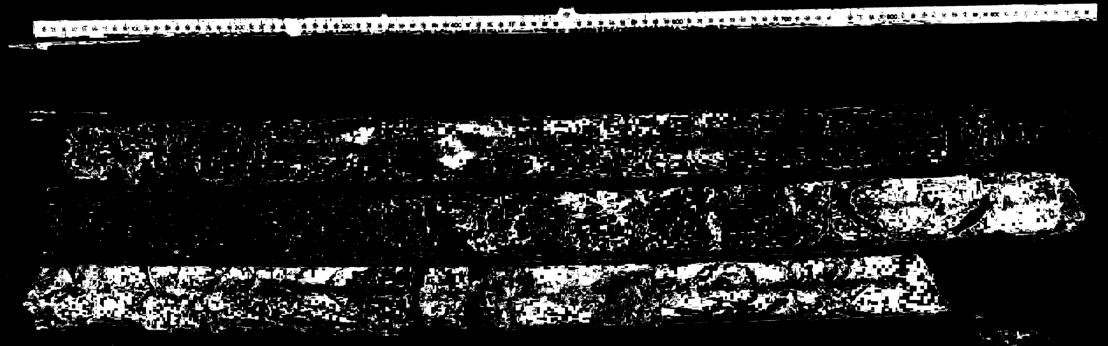}\\[4pt]
  \includegraphics[width=0.48\columnwidth]{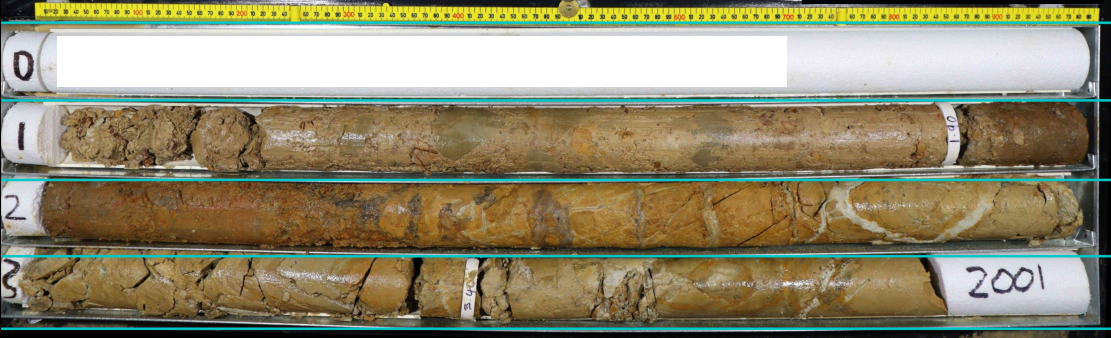}\hfill
  \includegraphics[width=0.48\columnwidth]{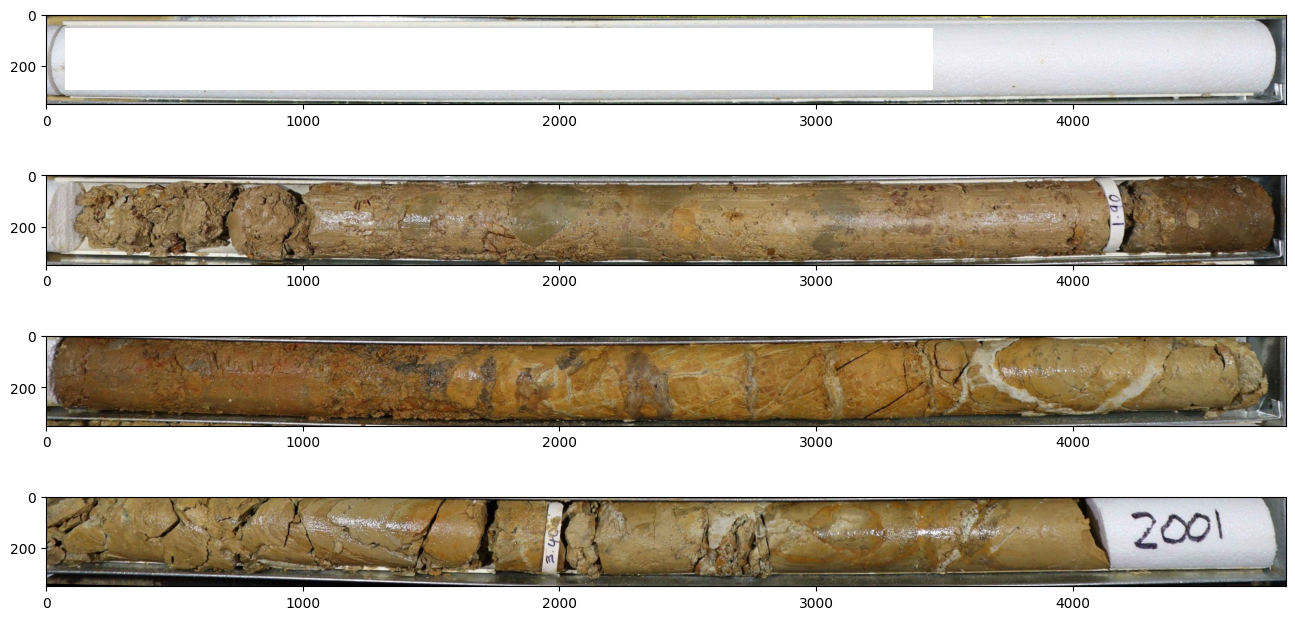}

  \caption{Core-row extraction. Top left: input core tray photograph. Top right:
  HSV mask used to detect the yellow ruler. Bottom left: candidate core-row
  boundaries. Bottom right: candidate row crops before manual quality control.
  Solid blocks redact client-identifying information. The first candidate contains
  the marker roll rather than core and is rejected during manual quality control.}
  \label{fig:extraction_pipeline}
\end{figure}

Each core tray image contains three to five core images, and each core image
spans approximately one metre of core. We use the ruler length to normalize
scale and convert pixel distances to physical distances. This conversion is
required for defect spacing categorization. Figure~\ref{fig:extraction_pipeline}
shows the extraction stages.

\subsection{Defect Spacing Label Extraction}
\label{sec:spacing_labels}

To obtain the report-derived labels used in the classification experiments, we
locate the ``SPACING'' column in each digital log report using the deterministic
procedure in Algorithm~\ref{alg:spacing_labels}. PyMuPDF reads words and their
bounding boxes from the PDF text layer, while vector drawing commands provide
the column rules; optical character recognition is not used. The vertical rules
on either side of the heading define the column bounds, and each adjacent pair
of rules defines a candidate region. Each region is rendered, converted to
grayscale, and binarized to identify the graphical spacing patches. Patch
heights are converted to report-depth intervals using the page scale and mapped
to the operator's spacing categories. The resulting report-derived labels are
aligned with core-row depth intervals and used as weak supervision in the
classification experiments.

\begin{algorithm}[htbp]
\caption{Defect Spacing Label Extraction}
\label{alg:spacing_labels}
\begin{algorithmic}[1]
\Require Page render $I_p$, column rules $L$, text-layer words $W$ with bounding boxes, top and bottom rules $T$ and $B$, and page number $p$
\Ensure Mapping from vertical regions to defect spacing labels
\State Locate ``SPACING'' in the text-layer words and obtain its bounding box
\If{the phrase is not found}
    \State Raise an error
\EndIf
\State Identify the vertical rules to the left and right of the phrase
\State Extract vertical rules within the horizontal and vertical column bounds
\State Sort these vertical lines to define column regions
\State Initialize a list of predefined defect spacing labels
\For{each adjacent pair of vertical lines}
    \State Extract the sub-image bounded by $T$, $B$, and the two vertical rules
    \State Convert to grayscale and binarize to highlight black patches
    \State Crop side margins to remove noise
    \State Calculate the pixel-to-length scale from the ruler height
    \State Detect connected black regions (spacing patches)
    \For{each patch}
        \State Convert its pixel height to real-world units using scaling
        \State Assign the corresponding defect spacing label to the region
    \EndFor
\EndFor
\State Remove any overlapping regions
\State \Return the mapping from physical regions to defect spacing labels
\end{algorithmic}
\end{algorithm}

\subsection{Segmentation Dataset}
\label{sec:seg_dataset}
We excluded tray positions in which no core was recovered and a white
placeholder label appeared instead of rock. These images contain no crack
evidence. Their bright, uniform surfaces would be scored as trivially correct
background and would inflate the metrics without demonstrating an ability to
segment rock. The remaining 5{,}087 core images form the dataset described
below.

For crack segmentation, we manually annotated the 5{,}087 core-row images with
polygonal masks. Each image has an extreme aspect ratio of approximately
$400\times5000$ pixels and contains thin, elongated cracks. Crack regions
touching white labels on the core surface were not annotated and were therefore
treated as background during training and evaluation. This choice avoids using
ambiguous label boundaries as crack supervision, but a prediction on a real
label-touching crack is counted as a false positive relative to the annotation. When adjacent cracks could not be visually separated, we drew one polygon around the complete cracked region. Annotations were saved in Labelme JSON format.

We randomly partitioned the boreholes into training, validation, and test sets
using a 70/15/15 ratio. All core-row images and derived crops from a borehole were assigned to the same partition. We also created a split-based variant by dividing each core-row image into ten vertical cells. Table~\ref{tab:dataset_splits} 
reports the resulting image and cell counts.

Data-owner restrictions prevent disclosure of the exact aggregate and
per-partition borehole counts; however, borehole identifiers were used
internally as grouping keys, and no borehole contributed images to more than one partition.

\begin{table}[pos=htbp]
\centering
\caption{Borehole-grouped partitions used for the crack-segmentation
experiments. All core-row images and derived crops from one borehole remain in
the same partition.}
\label{tab:dataset_splits}
\begin{tabular}{lccc}
\toprule
\textbf{Dataset unit} & \textbf{Train} & \textbf{Validation} & \textbf{Test} \\
\midrule
Core-row images & 3561 & 763 & 763 \\
Core splits & 35610 & 7630 & 7630 \\
\bottomrule
\end{tabular}
\end{table}

\section{Methods}

\begin{figure}[pos=tbp]
\centering
\resizebox{\columnwidth}{!}{%
\begin{tikzpicture}[
    node distance=0.6cm and 0.7cm,
    every node/.style={font=\scriptsize},
    block/.style={rectangle, draw=black, rounded corners=3pt,
                  minimum width=2.1cm, minimum height=1.3cm,
                  text centered, text width=2.0cm, inner sep=3pt},
    io/.style={block, fill=gray!20},
    proc/.style={block, fill=blue!10},
    output/.style={block, fill=green!10},
    arrow/.style={-{Latex[width=1.5mm]}, thick},
    lbl/.style={font=\tiny, midway, above, text=black}
]

% Pixel path: row 1
\node[io]     (input)    {Core Tray\\Image};
\node[proc,   right=of input]    (extract)  {Core Image\\Extraction};
\node[proc,   right=of extract]  (segment)  {Crack\\Segmentation\\(Gated UNet)};
\node[proc,   right=of segment]  (postproc) {Post-\\processing};
\node[output, right=of postproc] (spacing)  {Defect\\Spacing};

% Bedding branch: row 2
\node[proc,   below=1.0cm of segment] (bedding) {Bedding Angle\\Estimation\\(LSD + PCA)};
\node[output, right=of bedding]       (angles)  {Bedding\\Angles};

% Color branch: row 3
\node[proc,   below=0.6cm of bedding] (color)  {Color\\Detection\\(LAB space)};
\node[output, right=of color]         (colors) {Lithological\\Colors};

% Label path: row 4
\node[io] (report) at ($(input.center |- color.center) + (0,-2.3)$) {Log Report\\(PDF)};
\node[proc,   right=of report] (parse)  {Label Parsing\\(PDF text layer)};
\node[output, right=of parse]  (pseudo) {Spacing\\Pseudo-labels};

% Arrows: pixel path
\draw[arrow] (input)    -- node[lbl] {photo} (extract);
\draw[arrow] (extract)  -- node[lbl] {images} (segment);
\draw[arrow] (segment)  -- node[lbl] {masks}  (postproc);
\draw[arrow] (postproc) --                    (spacing);

% Arrows: branches from extraction
\draw[arrow] (extract.south) |- (bedding.west);
\draw[arrow] (extract.south) |- (color.west);
\draw[arrow] (bedding) -- (angles);
\draw[arrow] (color)   -- (colors);

% Arrows: label path
\draw[arrow] (report) -- node[lbl] {text} (parse);
\draw[arrow] (parse)  -- (pseudo);

% The two artifacts are matched, not derived from one another
\draw[{Latex[width=1.5mm]}-{Latex[width=1.5mm]}, densely dashed, gray]
      (input.south) -- (report.north)
      node[midway, right=2pt, font=\tiny, align=left, text=black]
      {matched by\\borehole + depth};

\end{tikzpicture}%
}
\caption{End-to-end system overview. The two input artifacts are independent and are matched through the borehole identifier and depth interval encoded in each filename. The core tray image provides all image pixels. Extracted core images pass to a gated UNet for crack segmentation. Post-processing converts the masks into defect spacing categories, while parallel branches estimate bedding angles and dominant lithological colors. The log report provides the reference labels. Its PDF text layer supplies defect spacing pseudo-labels for the self-supervised method in Section~\ref{sec:ssl} and the evaluation references used in Section~\ref{sec:results}.}
\label{fig:system_overview}
\end{figure}
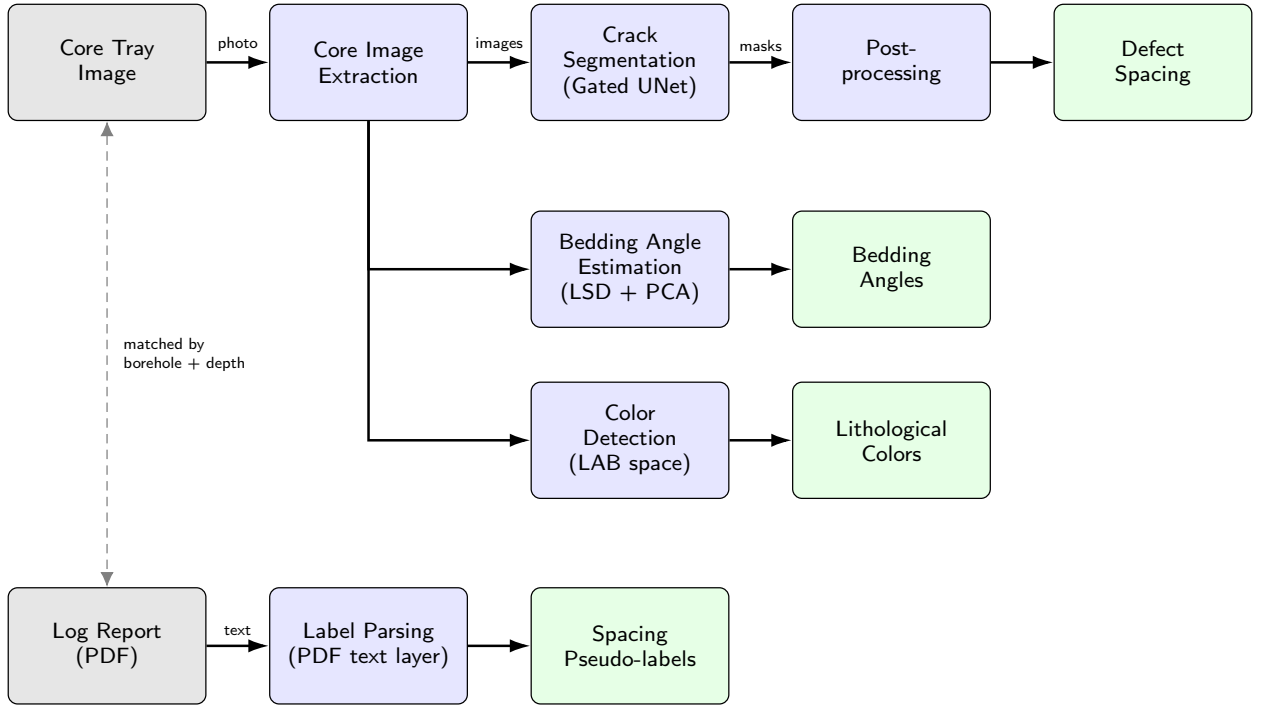

\subsection{Defect Spacing Classification}
\label{sec:ssl}
\textbf{Classical pipeline.} We apply glare removal, Canny edge detection, and
contour filtering to identify cracks. We then measure the crack-center
positions and the distances between them. The method detects many cracks but
also produces false positives under intensity variation and surface artifacts.

\begin{figure}[pos=tbp]
  \centering
  \includegraphics[width=0.48\columnwidth]{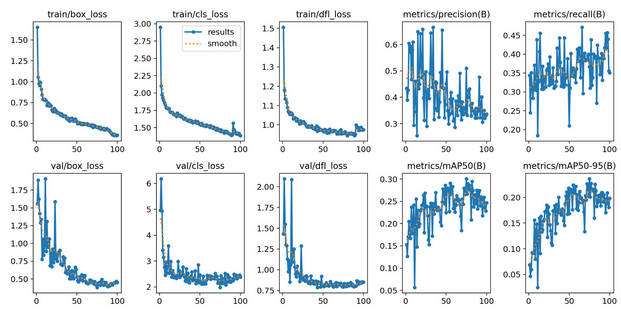}\hfill
  \includegraphics[width=0.48\columnwidth]{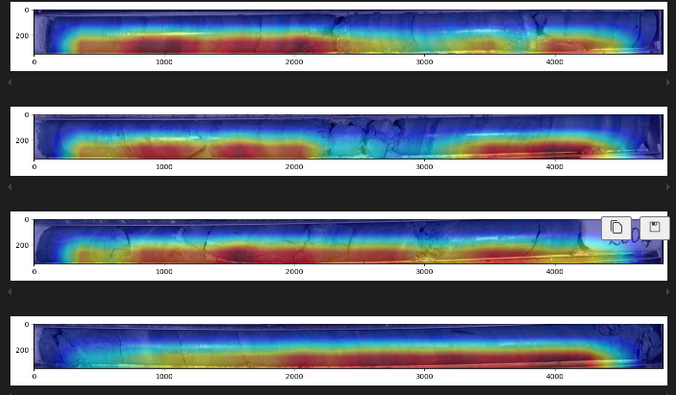}
  \caption{Supervised detection results. The YOLO training curves appear on the left. The Grad-CAM visualization on the right shows attention to image boundaries rather than crack regions.}
  \label{fig:yolo_training}
\end{figure}

\textbf{Supervised detection.} As a preliminary experiment separate from the
crack-segmentation dataset in Table~\ref{tab:dataset_splits}, we train a
YOLOv8-Large model on a 710-image bounding-box dataset for the six defect-spacing
classes (10, 30, 100, 300, 1000, and 3000). This dataset contains 546 training
images, 82 validation images, and 82 test images. Images are resized to
$640\times640$, and training runs for 100 epochs with a batch size of 32.
Although the overall mAP is reasonable, Grad-CAM reveals attention to irrelevant
image boundaries (Figure~\ref{fig:yolo_training}), indicating that the model learns spurious correlations.

\textbf{Self-supervised learning.}
We developed this method before pixel-level annotations became available. At
that stage, the log reports provided only defect-spacing categories, so we posed
the task as classification over core cells and evaluated representation learning
without annotated crack pixels. We report this classification branch as a
standalone experiment; its filtering procedure is not used to generate the
manual masks for the later segmentation experiments.

\textbf{Embedding-based label filtering.}
The report-derived categories (Section~\ref{sec:spacing_labels}) are noisy
because the logging geologist assigns a category to a depth interval by visual
assessment rather than by measuring every crack-to-crack distance. Aligning an
interval-level category with individual 100-mm core cells introduces additional
ambiguity. For each of the six spacing classes, we manually
verified 100 cells drawn only from the training boreholes and used their DINO
embeddings as a reference pool. Each remaining training cell was compared with
this pool using cosine similarity. If the majority class among its top-$5$
neighbors disagreed with its report-derived label, the cell was marked as
embedding-inconsistent and excluded from classifier training; it was not given
a replacement label. This procedure removed 5{,}723 of 35{,}610
training cells (16\%).

\textbf{Architecture.} We train a DINO ViT-Tiny encoder from scratch using only
unlabeled cells from the training boreholes and freeze it before training the
defect-spacing classifiers \citep{caron2021emerging}. Dividing each core image into 10 vertical segments
of fixed physical width provides a large pool of unlabeled samples. The DINO
encoder maps each sample to a 192-dimensional embedding, which passes to a
lightweight fully connected classifier. DINO uses the symmetrized cross-entropy
between student and teacher views:
\begin{equation}
\mathcal{L}_{\text{DINO}} = \frac{1}{2}\left[\mathcal{H}(t_1, s_2) + \mathcal{H}(t_2, s_1)\right],
\end{equation}
where $\mathcal{H}(t, s)$ is the cross-entropy between the sharpened teacher
output and the student prediction. The teacher logits are centered and
sharpened before softmax. An exponential moving average updates the teacher
weights.

A flat six-class classifier trained on the DINO embeddings reaches 68.6\%
overall accuracy. This aggregate result hides a clear error pattern. The model
reliably identifies the coarse, well-separated classes, achieving 93.8\% for
3000 and 78.1\% for 1000. In contrast, it confuses the adjacent fine classes,
achieving 46.9\% for 30 and 43.8\% for 100. The errors therefore arise from
confusion between neighboring bins rather than uniform task difficulty. This
finding motivates a decomposed decision process.

We therefore introduce a hierarchical inference pipeline
(Algorithm~\ref{alg:hierarchical}) that divides the six-class decision into
three binary stages. Each extracted core image spans approximately 1~m and is
divided into ten ordered cells of approximately 100~mm. The first stage labels
each cell as cracked or uncracked. Cracked cells pass through classifiers for
the 10, 30, and 100~mm categories, whereas uncracked cells are provisionally
assigned to the 300~mm category. We then scan the cells in spatial order. Cells are processed in continuous depth order, and an uncracked run at the end
of one core-row image is continued into the next depth-adjacent core-row image. Runs
of fewer than three consecutive uncracked cells retain the 300~mm category,
runs of three to ten uncracked cells are assigned to the 1000~mm category, and
runs of eleven or more uncracked cells are assigned to the open-ended 3000~mm category.
The coarser categories are therefore inferred from the lower bound provided by
the length of an uncracked run. Because the crack position within a 100-mm cell
is unknown, this run-length procedure provides an approximate category rather
than an exact crack-to-crack distance.

\begin{table}[pos=htbp]
\centering
\caption{Self-supervised defect spacing classification with DINO embeddings. The flat classifier predicts all six classes in one step. The hierarchical pipeline divides the same decision into three binary stages, each evaluated on the subset received from the previous stage.}
\label{tab:ssl_hierarchical}
\begin{tabular}{lcc}
\toprule
\textbf{Stage} & \textbf{Accuracy} & \textbf{F1} \\
\midrule
Flat six-class classifier & 0.686 & --- \\
\midrule
Stage 1: cracked vs.\ uncracked & 0.964 & 0.964 \\
Stage 2: 10 vs.\ all            & 0.946 & 0.946 \\
Stage 3: 30 vs.\ 100            & 0.801 & 0.805 \\
\bottomrule
\end{tabular}
\end{table}

\begin{algorithm}[htbp]
\caption{Hierarchical Spacing Inference}
\label{alg:hierarchical}
\begin{algorithmic}[1]
\Require Feature tensor $\mathbf{f}$ for cells ordered continuously across depth-adjacent core-row images
\Ensure Predicted labels $\hat{y}$, confidences $p$
\State $r_1 \leftarrow \text{CrackedUncrackedModel}(\mathbf{f})$
\State $(p_1, y_1) \leftarrow \text{Softmax}(r_1), \arg\max(r_1)$
\If{any $y_1 = 0$}
    \State $\mathbf{f}_1 \leftarrow \mathbf{f}[y_1 = 0]$
    \State $r_2 \leftarrow \text{TenVsAllModel}(\mathbf{f}_1)$
    \State $(p_2, y_2) \leftarrow \text{Softmax}(r_2), \arg\max(r_2)$
    \If{any $y_2 = 1$}
        \State $\mathbf{f}_2 \leftarrow \mathbf{f}_1[y_2 = 1]$
        \State $r_3 \leftarrow \text{ThirtyVsHundredModel}(\mathbf{f}_2)$
        \State $(p_3, y_3) \leftarrow \text{Softmax}(r_3), \arg\max(r_3)$
        \State Map $y_3$ labels as $0 \rightarrow 30$ and $1 \rightarrow 100$
        \State $y_2[y_2 = 1] \leftarrow y_3$, $p_2[y_2 = 1] \leftarrow p_3$
    \EndIf
    \State $y_2[y_2 = 0] \leftarrow 10$
    \State $y_1[y_1 = 0] \leftarrow y_2$, $p_1[y_1 = 0] \leftarrow p_2$
\EndIf
\State $y_1[y_1 = 1] \leftarrow 300$
\State $\hat{y} \leftarrow y_1$
\State Identify maximal runs $\mathcal{R}$ of consecutive labels $\hat{y}=300$ in the continuous depth order
\For{each run $R_j \in \mathcal{R}$ with length $r_j$}
    \If{$r_j \geq 11$}
        \State $\hat{y}[R_j] \leftarrow 3000$
    \ElsIf{$r_j \geq 3$}
        \State $\hat{y}[R_j] \leftarrow 1000$
    \EndIf
\EndFor
\State \Return $(p_1, \hat{y})$
\end{algorithmic}
\end{algorithm}

These results show that self-supervised representations support defect spacing
classification without pixel-level annotations. This capability is valuable
when such annotations are unavailable, as they were during this stage of the
work. Once annotations became available, however, we adopted segmentation as
the primary method because defect spacing is the \emph{distance between
successive cracks}. It is therefore a geometric property of crack positions
along the core axis.

Classification requires the network to infer this geometry implicitly and
return only a defect spacing category. The result omits the crack positions,
cannot represent multiple spacing regimes within one split, and is limited by
the split width. Segmentation instead identifies the cracks and measures the
intervals directly. A disputed defect spacing estimate can therefore be traced
to the mask that produced it. Classification also inherits the category
resolution of the log report, whereas segmentation measures a continuous
distance before assigning a category. We retain the self-supervised method as an annotation-free alternative for
defect-spacing classification.

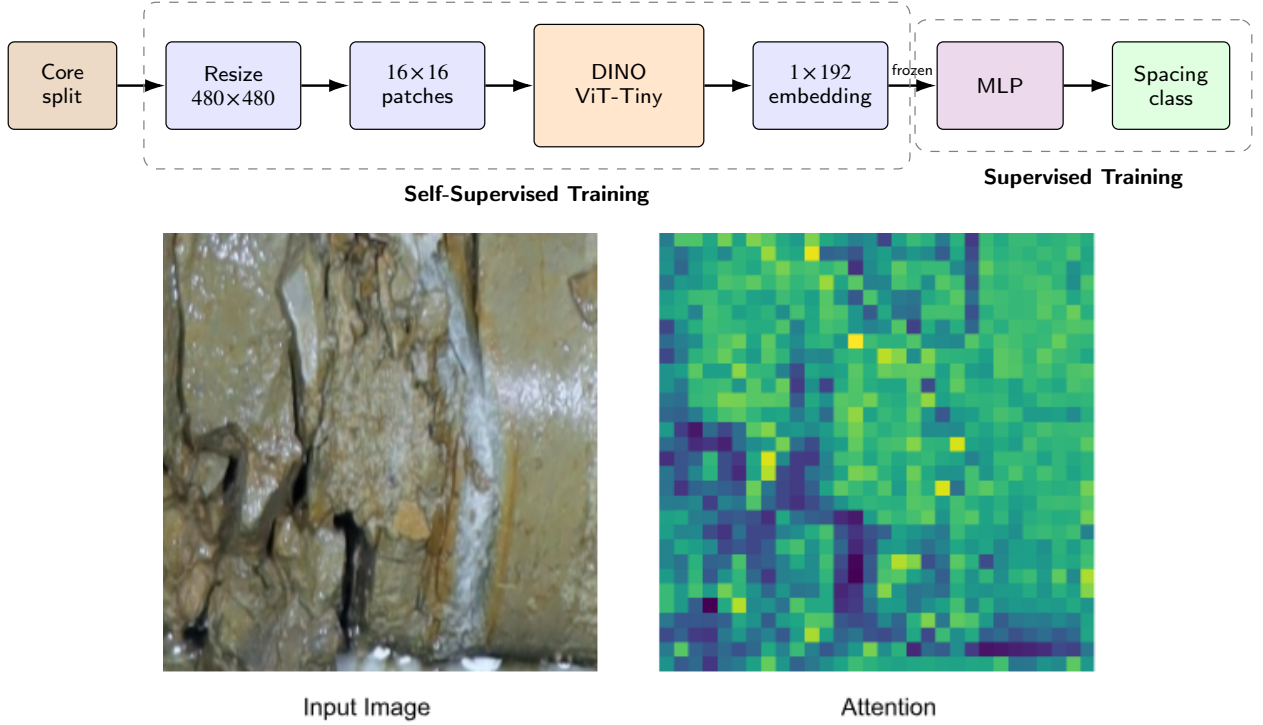
\begin{figure}[pos=tbp]
\centering
\resizebox{\columnwidth}{!}{%
\begin{tikzpicture}[
    node distance=0.55cm,
    every node/.style={font=\scriptsize},
    stage/.style={rectangle, draw=black, fill=blue!10, rounded corners=2pt,
                  minimum width=1.5cm, minimum height=1.0cm,
                  text centered, text width=1.4cm, inner sep=2pt},
    core/.style={rectangle, draw=black, fill=brown!30, rounded corners=2pt,
                 minimum width=1.2cm, minimum height=1.0cm,
                 text centered, text width=1.1cm, inner sep=2pt},
    dino/.style={rectangle, draw=black, fill=orange!20, rounded corners=2pt,
                 minimum width=1.9cm, minimum height=1.4cm,
                 text centered, text width=1.8cm, inner sep=2pt},
    mlp/.style={rectangle, draw=black, fill=violet!15, rounded corners=2pt,
                minimum width=1.4cm, minimum height=1.0cm,
                text centered, text width=1.3cm, inner sep=2pt},
    outp/.style={rectangle, draw=black, fill=green!12, rounded corners=2pt,
                 minimum width=1.3cm, minimum height=1.0cm,
                 text centered, text width=1.2cm, inner sep=2pt},
    grp/.style={draw=gray, dashed, rounded corners=4pt, inner sep=7pt},
    arr/.style={-{Latex[width=1.5mm]}, thick},
    title/.style={font=\scriptsize\bfseries, below=2pt}
]

\node[core]                      (split)  {Core\\split};
\node[stage, right=of split]     (resize) {Resize\\$480\!\times\!480$};
\node[stage, right=of resize]    (patch)  {$16\!\times\!16$\\patches};
\node[dino,  right=of patch]     (vit)    {DINO\\ViT-Tiny};
\node[stage, right=of vit]       (embed)  {$1\!\times\!192$\\embedding};
\node[mlp,   right=of embed]     (mlpnode){MLP};
\node[outp,  right=of mlpnode]   (out)    {Spacing\\class};

\draw[arr] (split)   -- (resize);
\draw[arr] (resize)  -- (patch);
\draw[arr] (patch)   -- (vit);
\draw[arr] (vit)     -- (embed);
\draw[arr] (embed)   -- node[font=\tiny, midway, above] {frozen} (mlpnode);
\draw[arr] (mlpnode) -- (out);

\node[grp, fit=(resize)(patch)(vit)(embed)] (sslbox) {};
\node[title] at (sslbox.south) {Self-Supervised Training};

\node[grp, fit=(mlpnode)(out)] (supbox) {};
\node[title] at (supbox.south) {Supervised Training};

\end{tikzpicture}%
}\\[6pt]
\includegraphics[width=0.75\columnwidth]{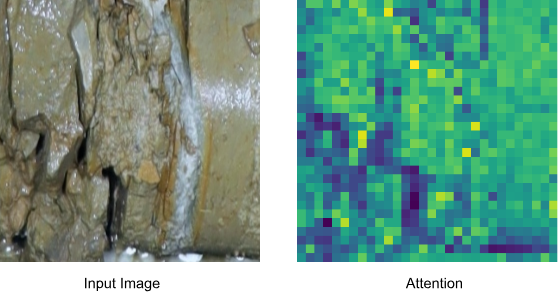}
\caption{Self-supervised defect-spacing classification. Each core cell is resized, divided into $16\times16$ patches, and encoded by a ViT-Tiny backbone trained without labels using only cells from the training boreholes. The frozen 192-dimensional embedding passes to a lightweight MLP trained on the retained report-derived labels after embedding-based filtering. The lower panel shows a representative attention map.}
\label{fig:dino_overview}
\end{figure}

\subsection{Crack Segmentation}
We evaluate YOLOv11-Nano, YOLOv11-Large, Mask R-CNN, and CrackCLIP on the
annotated dataset \citep{liang2025crackclip,radford2021learning}. We resize the
core images to $640\times640$ and use the borehole-grouped training, validation,
and test partitions in Table~\ref{tab:dataset_splits}. Both YOLOv11 variants are
trained for 100 epochs with a batch size of 16 and their default augmentations. We compare full-image training with a split-based variant in
which each core image is divided into 10 sub-images of $400\times500$ pixels.
This division expands the training set from 3{,}561 to 35{,}610 samples. The
full-image model outperforms the split-based model in both mAP@50 (0.76 versus
0.57) and precision (0.74 versus 0.58). The split crops therefore lose context
needed to distinguish cracks from surface artifacts. All subsequent
segmentation experiments use complete core images. Mask R-CNN is trained for
100 epochs with a batch size of 8 and an initial learning rate of
$1\times10^{-4}$. Its augmentations include horizontal flips and brightness
shifts.

\begin{figure}[pos=tbp]
  \centering
  \includegraphics[width=\columnwidth]{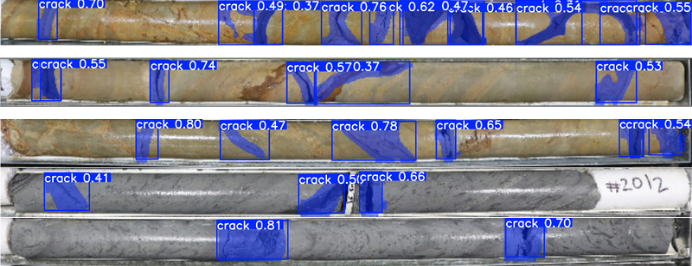}
  \caption{YOLOv11 instance segmentation on test core images. Bounding boxes, confidence scores, and blue masks show the predicted cracks. The model reliably detects high-contrast cracks in brown lithologies (top images). Predictions are sparser on dark gray core (bottom images), where the model misses several visible cracks.}
  \label{fig:yolo_seg_inference}
\end{figure}

Figure~\ref{fig:yolo_seg_inference} shows representative YOLOv11 predictions.
The detector identifies most distinct cracks with high confidence. However, it
misses low-contrast cracks in dark lithologies and divides some continuous
cracks into several detections. These errors motivate the fusion decoder below.

We combine multiple sources of evidence to improve segmentation. The following
sections first describe a multi-encoder UNet baseline and then present the gated
decoder that supersedes it.

\subsection{Multi-Encoder UNet Baseline}
\label{sec:unet3enc}
This baseline assigns one encoder to each of three inputs: the RGB core image,
the PiDiNet edge map, and the instance-segmentation mask. All three encoders use
ImageNet-pretrained ResNet34 weights. For each single-channel input, we
initialize the first convolution with the channel-wise mean of the pretrained
RGB kernels. The model fuses the three streams at all five encoder scales. At each scale, it
concatenates the feature maps and reduces them to one vector through global
average pooling. A two-layer MLP ($3C \rightarrow C/4 \rightarrow 3$) then
produces three softmax weights for their linear combination. A
$512\rightarrow1024$ bottleneck connects the encoders to five decoder blocks.
Each block upsamples through transposed convolution, concatenates the fused
skip connection at the corresponding scale, and applies a double convolution.
A final transposed convolution and $3\times3$ convolution produce the crack
probability map.

The gated decoder differs in the \emph{granularity} of its fusion. Global
pooling reduces the baseline's fusion weights to three scalars per core image.
The model can determine that edges are generally more reliable than masks for a
particular image, but it applies this decision uniformly to every pixel. It
cannot prefer the edge map in one region and the mask in another, even though
the sources fail at different locations within a core image.

\subsection{Spatially Gated U-Net Fusion}
\label{sec:gated_unet}
\begin{figure}[pos=tbp]
\centering
\resizebox{0.96\columnwidth}{!}{%
\begin{tikzpicture}[
    node distance=0.20cm,
    every node/.style={font=\scriptsize, align=center},
    layer/.style={rectangle, draw=black!65, line width=0.55pt,
                 rounded corners=1.5pt, fill=blue!12,
                 minimum width=2.55cm, minimum height=0.48cm,
                 inner sep=2pt},
    norm/.style={layer, fill=green!14},
    activation/.style={layer, fill=red!12},
    sigmoid/.style={layer, fill=orange!20},
    operation/.style={layer, fill=violet!14},
    group/.style={draw=black!35, line width=0.55pt,
                 rounded corners=3pt, inner sep=6pt},
    arr/.style={-{Latex[length=1.7mm,width=1.1mm]},
                line width=0.55pt, draw=black!65},
    title/.style={font=\scriptsize\bfseries, text=black!80}
]

% Attention gate: one sequential path
\node[layer] (g1) {Conv2d: $2\!\to\!16$};
\node[norm, below=of g1] (g2) {BatchNorm$(16)$};
\node[activation, below=of g2] (g3) {ReLU};
\node[layer, below=of g3] (g4) {Conv2d: $16\!\to\!8$};
\node[norm, below=of g4] (g5) {BatchNorm$(8)$};
\node[activation, below=of g5] (g6) {ReLU};
\node[layer, below=of g6] (g7) {Conv2d: $8\!\to\!1$};
\node[sigmoid, below=of g7] (g8) {Sigmoid};
\foreach \a/\b in {g1/g2,g2/g3,g3/g4,g4/g5,g5/g6,g6/g7,g7/g8}
  \draw[arr] (\a) -- (\b);
\node[group, fit=(g1)(g8)] (gatebox) {};
\node[title, above=0.12cm of gatebox] {Attention gate};

% ConvBlock
\node[layer, right=1.35cm of g1] (c1) {Conv2d};
\node[norm, below=of c1] (c2) {BatchNorm + ReLU};
\node[layer, below=of c2] (c3) {Conv2d};
\node[norm, below=of c3] (c4) {BatchNorm + ReLU};
\draw[arr] (c1) -- (c2);
\draw[arr] (c2) -- (c3);
\draw[arr] (c3) -- (c4);
\node[group, fit=(c1)(c4)] (convbox) {};
\node[title, above=0.12cm of convbox] {ConvBlock};

% Down component
\node[operation, right=1.35cm of c1] (p1) {MaxPool $2\!\times\!2$};
\node[layer, below=of p1] (p2) {ConvBlock};
\draw[arr] (p1) -- (p2);
\node[group, fit=(p1)(p2)] (downbox) {};
\node[title, above=0.12cm of downbox] {Down};

% Up component
\node[operation, below=1.15cm of p2] (u1) {Upsample $\times2$};
\node[layer, below=of u1] (u2) {ConvBlock};
\draw[arr] (u1) -- (u2);
\node[group, fit=(u1)(u2)] (upbox) {};
\node[title, above=0.12cm of upbox] {Up};

\end{tikzpicture}%
}
\caption{Components of the spatially gated U-Net. The attention gate maps the
concatenated two-channel input through convolutional outputs of 16, 8, and 1
channel, using batch normalization and ReLU after the first two convolutions
and sigmoid after the last. ConvBlock applies two convolution--normalization--
activation stages. Down applies max pooling before a ConvBlock, whereas Up
applies twofold upsampling before a ConvBlock.}
\label{fig:unet_components}
\end{figure}
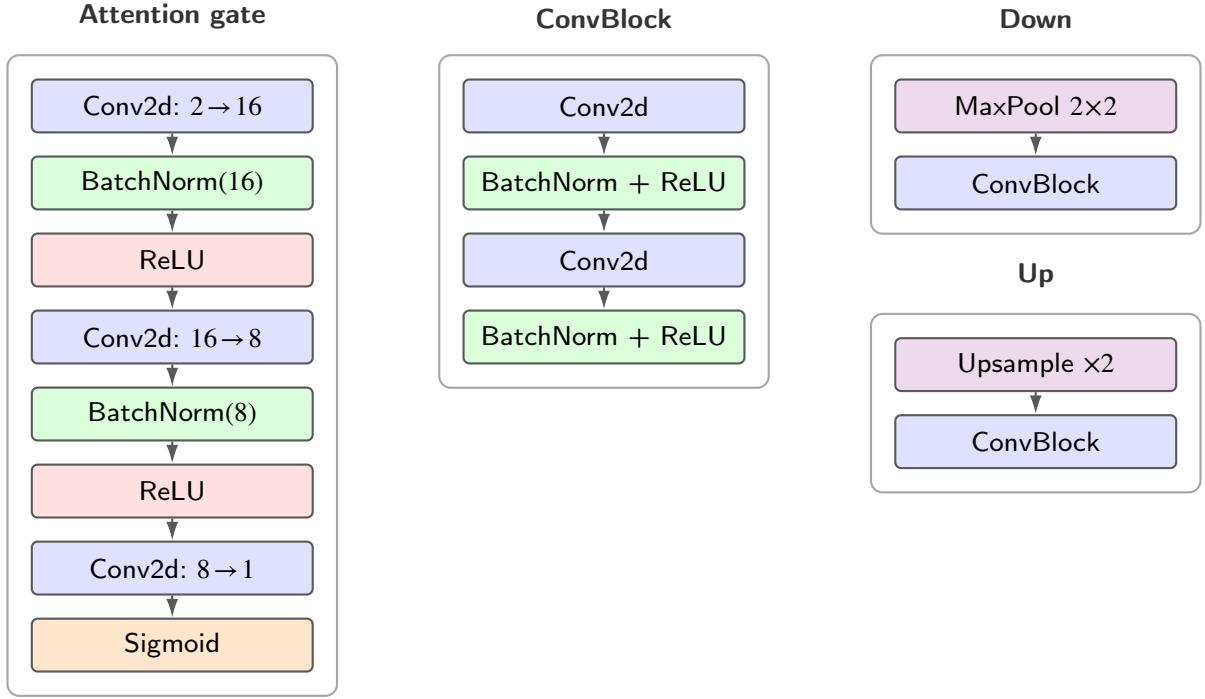

The model receives two binary single-channel inputs: a segmentation mask $S$
from Mask R-CNN or CrackCLIP and an edge map $E$ from PiDiNet. The inputs are
concatenated and passed through a sequential attention gate. Its three
convolutions reduce the channel dimension from 2 to 16, from 16 to 8, and from
8 to 1. Batch normalization and ReLU follow the first two convolutions, and a
sigmoid produces the one-channel gated representation supplied to the U-Net
(Figure~\ref{fig:unet_components}). This learned representation permits the
relative contribution of the two input sources to vary spatially.

The gated output passes to an initial 32-channel ConvBlock and then through
three encoder Down blocks with 64, 128, and 256 output channels. A 256-channel
ConvBlock forms the bottom stage. Four decoder Up blocks produce 128, 64, 32,
and 32 channels and combine the decoder representation with encoder skip
features. 
A final $1\times1$ convolution produces one crack-logit map at the
input spatial resolution.
We train separate gated models for the Mask R-CNN and CrackCLIP segmentation
inputs. All input and reference masks are resized to $512\times512$ pixels.
Training uses Adam with a learning rate of $10^{-4}$, a batch size of 32, and
100 epochs. Flips and rotations are applied jointly to the segmentation mask,
edge map, and reference mask. The objective is the unweighted sum of binary
cross-entropy and soft Dice losses. We retain the checkpoint selected on the
validation partition.

The edge map, segmentation mask, and attention map can each be inspected
independently. These intermediate outputs make the pipeline more interpretable
than single-encoder alternatives.

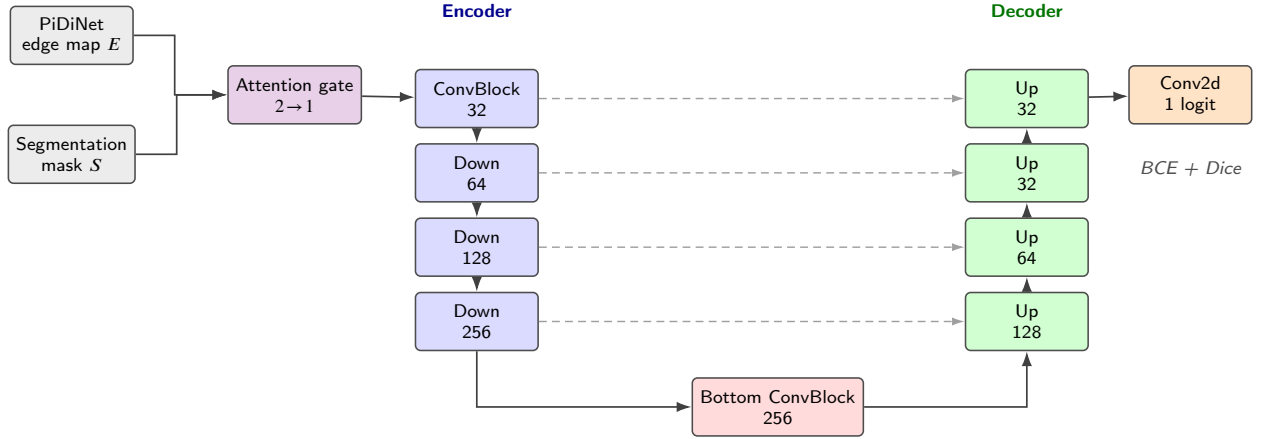
\begin{figure}[pos=tbp]
\centering
\resizebox{\columnwidth}{!}{%
\begin{tikzpicture}[
    every node/.style={font=\scriptsize, align=center},
    block/.style={rectangle, draw=black!70, line width=0.6pt,
                 rounded corners=2pt, minimum width=1.65cm,
                 minimum height=0.78cm, inner sep=3pt},
    input/.style={block, fill=gray!15},
    gate/.style={block, fill=violet!18},
    enc/.style={block, fill=blue!14},
    bottleneck/.style={block, fill=red!14},
    dec/.style={block, fill=green!18},
    output/.style={block, fill=orange!22},
    arr/.style={-{Latex[length=2.0mm,width=1.4mm]},
                line width=0.65pt, draw=black!75},
    skip/.style={-{Latex[length=1.7mm,width=1.1mm]},
                 line width=0.55pt, densely dashed, draw=gray!75}
]

% Inputs and spatial gate
\node[input] (edge) at (0,2.85) {PiDiNet\\edge map $E$};
\node[input] (seg)  at (0,1.25) {Segmentation\\mask $S$};
\node[gate]  (gate) at (3.00,2.05) {Attention gate\\$2\!\to\!1$};

% Encoder path
\node[enc] (e1) at (5.45,2.00) {ConvBlock\\32};
\node[enc] (e2) at (5.45,1.00) {Down\\64};
\node[enc] (e3) at (5.45,0.) {Down\\128};
\node[enc] (e4) at (5.45,-1.00) {Down\\256};
\node[bottleneck] (bn) at (9.5,-2.15) {Bottom ConvBlock\\256};

% Decoder path
\node[dec] (d4) at (12.85,-1.) {Up\\128};
\node[dec] (d3) at (12.85,0) {Up\\64};
\node[dec] (d2) at (12.85,1.00) {Up\\32};
\node[dec] (d1) at (12.85,2.00) {Up\\32};
\node[output] (out) at (15.05,2.05) {Conv2d\\1 logit};

% Stage headings
\node[font=\scriptsize\bfseries, text=blue!55!black]
      at (5.45,3.18) {Encoder};
\node[font=\scriptsize\bfseries, text=green!45!black]
      at (12.85,3.18) {Decoder};

% Input paths
\draw[arr] (edge.east) -- ++(0.55,0) |- (gate.west);
\draw[arr] (seg.east)  -- ++(0.55,0) |- (gate.west);

% Encoder and bottom paths
\draw[arr] (gate) -- (e1);
\draw[arr] (e1) -- (e2);
\draw[arr] (e2) -- (e3);
\draw[arr] (e3) -- (e4);
\draw[arr] (e4.south) |- (bn.west);

% Decoder path
\draw[arr] (bn.east) -| (d4.south);
\draw[arr] (d4.north) -- (d3.south);
\draw[arr] (d3.north) -- (d2.south);
\draw[arr] (d2.north) -- (d1.south);
\draw[arr] (d1) -- (out);

% Encoder--decoder skip connections
\draw[skip] (e4.east) -- (d4.west);
\draw[skip] (e3.east) -- (d3.west);
\draw[skip] (e2.east) -- (d2.west);
\draw[skip] (e1.east) -- (d1.west);

% Training annotation
\node[font=\scriptsize\itshape, text=black!70, below=0.38cm of out]
      {BCE + Dice};
\end{tikzpicture}%
}
\caption{Spatially gated U-Net architecture. The attention gate combines a
PiDiNet edge map and a segmentation mask into a one-channel representation.
The encoder uses an initial 32-channel ConvBlock followed by Down blocks with
64, 128, and 256 channels. The decoder uses Up blocks with 128, 64, 32, and 32
channels; dashed arrows denote encoder--decoder skip connections. A final
convolution produces the crack-logit map.}
\label{fig:gate_fusion_arch}
\end{figure}

\subsection{Post-processing for Spacing}
\begin{figure}[pos=tbp]
  \centering
  \includegraphics[width=\columnwidth]{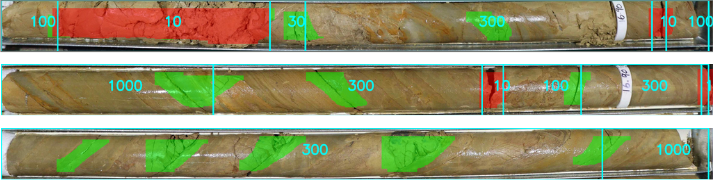}
  \caption{Defect spacing recovered from three core images. Predicted crack segments are green, and broader cracked regions are red. The gaps between consecutive segments define the cyan interval boundaries and their defect spacing categories. Dense groups of cracks merge into cracked regions and receive the finest categories (e.g., 10), whereas intact intervals between isolated cracks receive coarser categories (e.g., 300 or 1000).}
  \label{fig:post_processing_results}
\end{figure}

We convert each segmentation mask into crack segments and broader cracked
regions and order them along the core axis. For segment $i$, let $l_i$ and $r_i$
denote its left (start) and right (end) bounds, respectively. The gap between
successive non-overlapping segments $i-1$ and $i$ is
\begin{equation}
d_i=l_i-r_{i-1}.
\end{equation}
We map these distances to fixed defect spacing ranges
(Table~\ref{tab:distance_label_mapping}) to produce structured annotations for
the core interval. The ranges correspond to the geotechnical categories used in
the log reports and therefore allow direct comparison with the recorded
annotations. Figure~\ref{fig:post_processing_results} shows representative
outputs for three core images.

\subsection{Bedding Angle Estimation}
We estimate bedding angles by detecting line segments with LSD, removing
near-horizontal artifacts, clustering collinear segments, and fitting line
models with PCA. Each detected segment
$\ell_i = (x_{i0}, y_{i0}, x_{i1}, y_{i1})$ has orientation
$\theta_i = \arctan2(y_{i1} - y_{i0}, x_{i1} - x_{i0})$. We suppress
near-horizontal segments and form collinear clusters subject to angular,
perpendicular-distance, and gap constraints. PCA then determines the dominant
direction of each cluster.

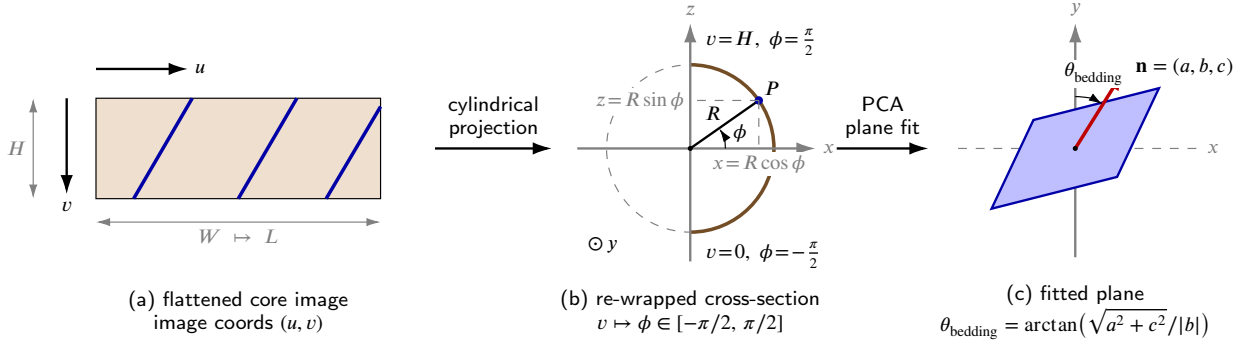
\begin{figure}[pos=tbp]
\centering
\resizebox{\columnwidth}{!}{%
\begin{tikzpicture}[
    every node/.style={font=\scriptsize},
    arr/.style={-{Latex[width=1.5mm]}, thick},
    trace/.style={very thick, blue!65!black},
    dim/.style={{Latex[width=1mm]}-{Latex[width=1mm]}, gray}
]

%% ---------- (a) flattened core image ----------
\begin{scope}[local bounding box=P1]
  \fill[brown!25, draw=black] (0,0) rectangle (3.4,1.2);
  \draw[trace] (0.45,0) -- (1.15,1.2);
  \draw[trace] (1.70,0) -- (2.40,1.2);
  \draw[trace] (2.75,0) -- (3.40,1.1);
  % axes
  \draw[arr] (0,1.55) -- (1.1,1.55) node[right=-1pt] {$u$};
  \draw[arr] (-0.35,1.2) -- (-0.35,0.05) node[below=-1pt] {$v$};
  % extents
  \draw[dim] (0,-0.28) -- (3.4,-0.28) node[midway, below=-1pt] {$W \;\mapsto\; L$};
  \draw[dim] (-0.75,0) -- (-0.75,1.2) node[midway, left=-1pt] {$H$};
\end{scope}
\node[align=center] at (1.7,-1.35) {(a) flattened core image\\image coords $(u,v)$};

%% ---------- (b) cross-section ----------
\begin{scope}[shift={(7.1,0.6)}, local bounding box=P2]
  \draw[dashed, gray] (0,0) circle (1.0);
  % photographed half-circumference
  \draw[very thick, brown!60!black] (-90:1.0) arc (-90:90:1.0);
  % axes
  \draw[arr, gray] (-1.35,0) -- (1.5,0) node[right=-1pt] {$x$};
  \draw[arr, gray] (0,-1.35) -- (0,1.5) node[above=-1pt] {$z$};
  \node at (-1.05,-1.15) {$\odot\,y$};
  % point P
  \coordinate (Pp) at (35:1.0);
  \draw[thick] (0,0) -- (Pp) node[midway, above left=-3pt] {$R$};
  \fill[blue!65!black] (Pp) circle (0.05);
  \node[above right=-2pt] at (Pp) {$P$};
  % angle phi
  \draw[-{Latex[width=1mm]}] (0.42,0) arc (0:35:0.42);
  \node at (17:0.62) {$\phi$};
  % projections
  \draw[dashed, gray] (Pp) -- ($(0,0)!(Pp)!(1,0)$) node[below=1pt, fill=white, inner sep=1pt] {$x\!=\!R\cos\phi$};
  \draw[dashed, gray] (Pp) -- ($(0,0)!(Pp)!(0,1)$) node[left=1pt, fill=white, inner sep=1pt] {$z\!=\!R\sin\phi$};
  % endpoints of the visible arc
  \node[above right=0pt and 2pt, align=left] at (90:1.0) {$v\!=\!H,\; \phi\!=\!\tfrac{\pi}{2}$};
  \node[below right=0pt and 2pt, align=left] at (-90:1.0) {$v\!=\!0,\; \phi\!=\!-\tfrac{\pi}{2}$};
  \fill (0,0) circle (0.03);
\end{scope}
\node[align=center] at (7.1,-1.35) {(b) re-wrapped cross-section\\$v \mapsto \phi \in [-\pi/2,\, \pi/2]$};

%% ---------- (c) plane fit and bedding angle ----------
\begin{scope}[shift={(11.7,0.6)}, local bounding box=P3]
  \draw[dashed, gray] (-1.4,0) -- (1.5,0) node[right=-1pt] {$x$};
  \draw[arr, gray] (0,-1.3) -- (0,1.5) node[above=-1pt] {$y$};
  % fitted bedding plane
  \fill[blue!25, draw=blue!65!black, thick]
      (-1.0,-0.72) -- (0.5,-0.34) -- (1.0,0.72) -- (-0.5,0.34) -- cycle;
  % plane normal and its angle to the longitudinal core axis
  \draw[arr, red!75!black, very thick] (0,0) -- (58:1.15);
  \node[right=0pt, align=left] at (58:1.15) {$\mathbf{n}=(a,b,c)$};
  \draw[-{Latex[width=1mm]}] (0,0.62) arc (90:58:0.62);
  \node[fill=white, inner sep=1pt] at (76:0.92) {$\theta_{\mathrm{bedding}}$};
  \fill (0,0) circle (0.03);
\end{scope}
\node[align=center] at (11.7,-1.35) {(c) fitted plane\\$\theta_{\mathrm{bedding}}=\arctan\!\big(\sqrt{a^2+c^2}/|b|\big)$};

%% ---------- stage arrows ----------
\draw[arr] (4.05,0.6) -- (5.45,0.6) node[midway, above, align=center] {cylindrical\\projection};
\draw[arr] (8.85,0.6) -- (9.95,0.6) node[midway, above, align=center] {PCA\\plane fit};

\end{tikzpicture}%
}
\caption{Geometry of bedding-angle estimation. (a) The long image coordinate $u$ follows core length $L$, while $v$ spans the visible diameter. (b) Each trace point is projected onto the photographed half of the cylindrical surface: $v$ determines $\phi$, $(x,z)$ span the circular cross-section, and the longitudinal axis $y$ points out of the page. (c) PCA fits a plane to the projected 3D trace points. The acute angle between its normal $\mathbf{n}$ and the longitudinal core axis $y$ is the reported bedding angle. Fragmented core violates the single-cylinder assumption and causes the principal failure mode shown in Figure~\ref{fig:bedding_angle}.}
\label{fig:bedding_geometry}
\end{figure}

We project the clustered trace points onto a cylindrical core model
(Figure~\ref{fig:bedding_geometry}). A line orientation measured only in the
2D image plane is insufficient to determine the 3D bedding-plane normal.
Therefore, we re-wrap the detected trace onto the photographed half of the
cylindrical surface before fitting a plane. The long image coordinate $u$
follows the longitudinal core axis, which we denote by $y$, whereas $v$ spans
the visible diameter and parameterizes the photographed half-circumference. For
image coordinates $(u,v)$, the cylindrical projection is
\begin{equation}
y=\frac{u}{W}L, \qquad
\phi=\frac{v}{H}\pi-\frac{\pi}{2}, \qquad
x=R\cos\phi, \qquad z=R\sin\phi.
\end{equation}
Thus, $y$ is the longitudinal core-axis coordinate and $(x,z)$ span the circular
cross-section. If the fitted plane normal is $\mathbf{n}=(a,b,c)$ in $(x,y,z)$
coordinates, we define the bedding angle as the acute normal-to-axis angle
\begin{equation}
\theta_{\mathrm{bedding}}
=\arctan\!\left(\frac{\sqrt{a^2+c^2}}{|b|}\right).
\end{equation}
A plane perpendicular to the core axis has a normal parallel to $y$ and gives $\theta_{\mathrm{bedding}}=0^\circ$, whereas more oblique planes give larger angles. We report bedding angles in the $10^\circ$ categories used in core logging. The log-report references use the same definition and resolution, which permits the comparison in Table~\ref{tab:bedding_angle_evaluations}.

\begin{figure}[pos=tbp]
  \centering
  \includegraphics[width=\columnwidth]{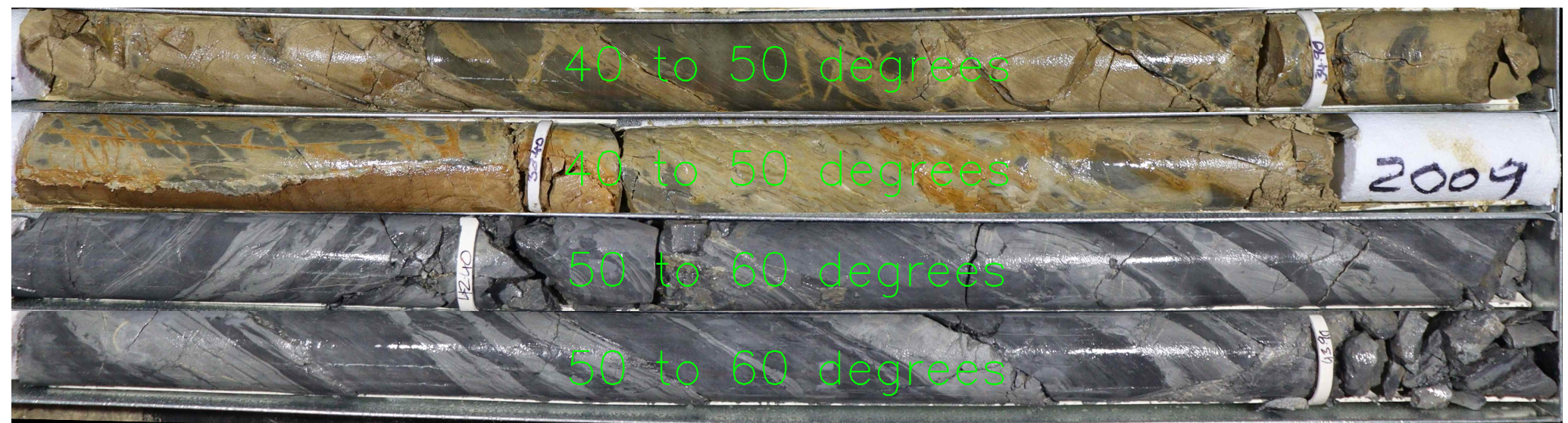}
  \caption{Bedding angle estimates for four core images. The upper two images show brown lithologies with predicted angles of $40$--$50^\circ$. The lower two show dark gray core with steeper bedding angles of $50$--$60^\circ$. Estimates are stable for intact core but degrade under fragmentation (lower right), which violates the cylindrical geometry assumption.}
  \label{fig:bedding_angle}
\end{figure}

\subsection{Lithological Color Detection}
\begin{figure}[pos=tbp]
  \centering
  \includegraphics[width=\columnwidth]{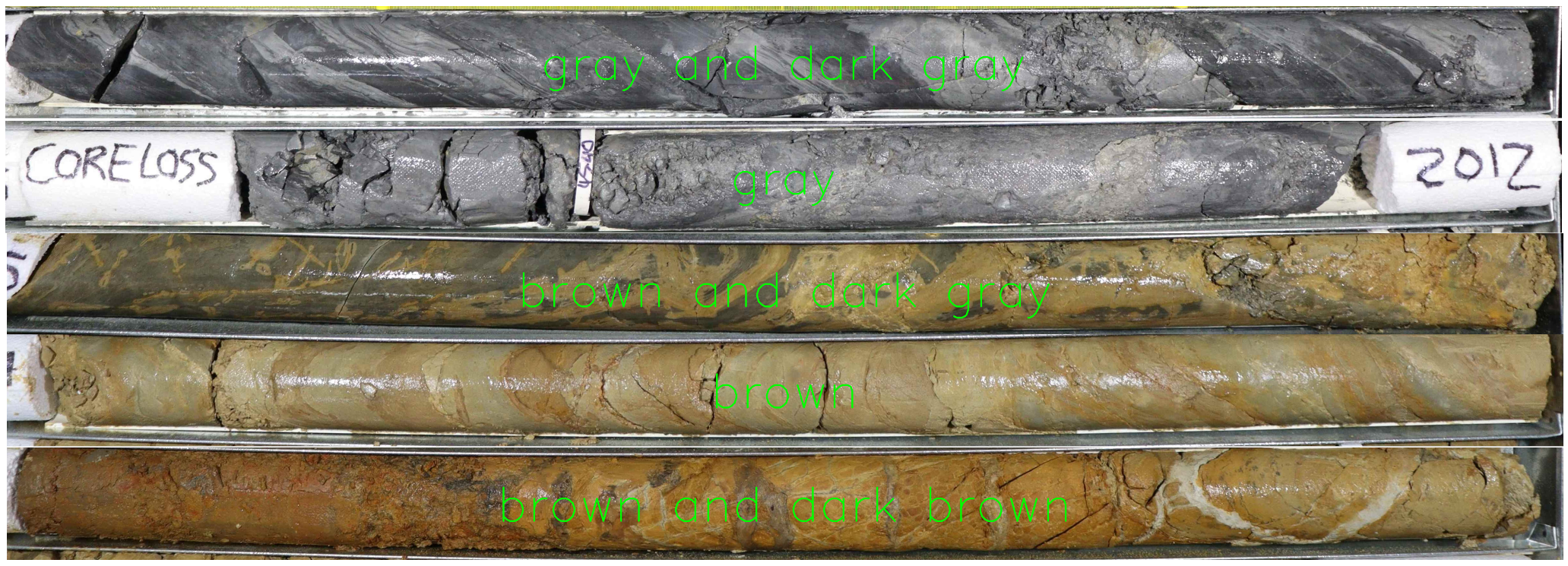}
  \caption{Lithological color descriptors for five core images. The method returns one dominant color (``gray'' or ``brown'') or a composite descriptor when two colors persist across consecutive core splits (``gray and dark gray,'' ``brown and dark gray,'' or ``brown and dark brown''). The near-white mask suppresses core-loss markers in the second image and excludes them from aggregation.}
  \label{fig:coloring}
\end{figure}

We divide each core image into vertical core splits and convert them to LAB
color space. The mean color summarizes each split, and the nearest reference
assigns its semantic color. A mask removes near-white regions to prevent core
loss from biasing the result. We retain only colors that occur in consecutive
splits and order the final descriptors by a fixed geological precedence. Let
$\bar{\mathbf{c}}_k$ denote the mean LAB color of split $k$, let
$\mathbf{r}_i$ denote reference color $i$, and let $\gamma_i$ denote its
semantic label. We assign
\begin{equation}
i_k^{\star}=\arg\min_i
\left\|\bar{\mathbf{c}}_k-\mathbf{r}_i\right\|_2,
\qquad
\hat{\gamma}_k=\gamma_{i_k^{\star}}.
\end{equation}
We discard core splits with a high proportion of white pixels and retain only
colors found in at least two consecutive splits. This procedure stabilizes the
labels under illumination variation, surface artifacts, and core loss.
Figure~\ref{fig:coloring} shows representative descriptors.

\section{Experiments and Results}
\label{sec:results}

We evaluate all segmentation models on the same 763 borehole-held-out test
images. If an upstream model produces no detection for an image, we retain that
image and represent the missing prediction by an empty mask. Probability
outputs are thresholded at 0.5. True-positive, false-positive, and
false-negative pixels are then accumulated over the complete test set for the
crack class. We report crack-class intersection over union,
\begin{equation}
\operatorname{IoU}_{\mathrm{crack}}
=\frac{TP}{TP+FP+FN},
\end{equation}
precision $TP/(TP+FP)$, recall $TP/(TP+FN)$, and
$F_1=2TP/(2TP+FP+FN)$. Thus, the reported metrics describe one fixed binary
operating point computed from aggregate test-set pixel counts.

The figures show representative outputs, and
Tables~\ref{tab:segmentation_results_comparison},
\ref{tab:bedding_angle_evaluations}, and \ref{tab:coloring_evaluation}
summarize the quantitative results. The main text focuses on the strongest
segmentation models and uses the remaining baselines as supporting evidence.
The Mask R-CNN gated-fusion configuration outperforms the detector-only
baselines and produces the highest crack-mask result among the evaluated
models. The self-supervised classifier remains useful when pixel annotations
are unavailable, but its decisions are less spatially transparent. We therefore
treat it as a complementary method.

The gated U-Net with Mask R-CNN input achieves the highest crack-class IoU and
F1. Its binary masks preserve many thin crack boundaries while suppressing
surface artifacts in the representative examples. The masks are subsequently
passed to the deterministic spacing procedure illustrated in Figure~\ref{fig:post_processing_results}; this figure is qualitative and does not constitute a downstream spacing-accuracy evaluation.

\textbf{Bedding angle and lithological color.} Both estimators are
training-free and apply fixed geometric or colorimetric rules. No data is used
for fitting, so every core image remains available for evaluation. We evaluate
each method on 1{,}200 core images sampled randomly from the corpus. The parser
used for defect spacing also extracts the corresponding bedding angle and
color references from the log reports (Section~\ref{sec:spacing_labels}). The
results in Tables~\ref{tab:bedding_angle_evaluations} and
\ref{tab:coloring_evaluation} therefore measure \emph{agreement with the log
report}, not correctness against an independent geological measurement. The
references are subject to the logging variability discussed in
Section~\ref{sec:intro_problem}. These results indicate how closely the methods
reproduce a geologist's recorded observations rather than their absolute
physical accuracy.

Bedding angle reconstruction agrees with the recorded angle category for
75.4\% of the core images (Table~\ref{tab:bedding_angle_evaluations}). The
estimated angles also follow the dominant structural trends visible in the core
(Figure~\ref{fig:bedding_angle}). Lithological color agrees with the log report
for 84.7\% of the core images (Table~\ref{tab:coloring_evaluation} and
Figure~\ref{fig:coloring}). This higher agreement is expected because color is
a low-frequency property of the complete image and remains stable under
LAB-space averaging. Bedding angle estimation instead depends on line evidence
that fragmentation can destroy. Most color errors occur under extreme
illumination or extensive core loss, whereas most bedding angle errors occur in
fragmented core that violates the cylindrical assumption.

% Figures moved to their respective sections

\begin{table}[pos=htbp]
\centering
\caption{Mapping of distance ranges to defect spacing labels.}
\label{tab:distance_label_mapping}
\begin{tabular}{lc}
\toprule
\textbf{Distance ($x$)} & \textbf{Label} \\
\midrule
$x \leq 10$ & 10 \\
$10 < x \leq 30$ & 30 \\
$30 < x \leq 100$ & 100 \\
$100 < x \leq 300$ & 300 \\
$300 < x \leq 1000$ & 1000 \\
$1000 < x$ & 3000 \\
\bottomrule
\end{tabular}
\end{table}

\begin{table}[pos=htbp]
\centering
\caption{Quantitative comparison of crack-segmentation models. Crack IoU, precision, recall, and F1 are computed pixel-wise for the crack class from the final binary masks.}
\label{tab:segmentation_results_comparison}
\begin{tabular}{lcccc}
\toprule
Model & Crack IoU & Precision & Recall & F1 \\
\midrule
YOLOv11 Nano & 0.2615 & 0.2745 & 0.8465 & 0.4146 \\
YOLOv11 Large & 0.2746 & 0.2967 & 0.7867 & 0.4309 \\
Mask R-CNN & 0.2717 & 0.2796 & 0.9056 & 0.4273 \\
CrackCLIP & 0.3000 & 0.4300 & 0.5000 & 0.4600 \\
\midrule
UNet ResNet34 (3 Encoders) & 0.6394 & 0.6884 & 0.8998 & 0.7801 \\
UNet Gate Fusion (CrackCLIP) & 0.2149 & 0.3440 & 0.3639 & 0.3537 \\
UNet Gate Fusion (Mask R-CNN) & \textbf{0.7539} & \textbf{0.8160} & \textbf{0.9083} & \textbf{0.8597} \\
\bottomrule
\end{tabular}
\end{table}

\begin{table}[pos=htbp]
\centering
\caption{Bedding angle estimation results using LSD-based line detection with cylindrical PCA fitting.}
\label{tab:bedding_angle_evaluations}
\begin{tabular}{lccccc}
\toprule
\textbf{Method} & Accuracy & Precision & Recall & F1 Score & Test Size \\
\midrule
LSD + Cylindrical PCA & 0.7538 & 0.7684 & 0.7421 & 0.7550 & 1200 \\
\bottomrule
\end{tabular}
\end{table}

\begin{table}[pos=htbp]
\centering
\caption{Lithological color detection results using LAB-space nearest-reference classification.}
\label{tab:coloring_evaluation}
\begin{tabular}{lccccc}
\toprule
\textbf{Method} & Accuracy & Precision & Recall & F1 Score & Test Size \\
\midrule
LAB Nearest-Reference & 0.8467 & 0.8523 & 0.8416 & 0.8469 & 1200 \\
\bottomrule
\end{tabular}
\end{table}

\section{Discussion}
The experiments evaluate two complementary approaches to defect-spacing
analysis. The weakly supervised branch predicts report-derived spacing
categories without crack masks, whereas the supervised branch first localizes
cracks and then derives spacing through deterministic post-processing. The
gated segmentation configuration achieves the highest crack-mask performance
among the evaluated models. Figure~\ref{fig:post_processing_results}
qualitatively illustrates the spacing output obtained from its predicted masks. 
We did not quantitatively evaluate the resulting interval distances or spacing categories against spacing derived from manual masks or against independent geological measurements; downstream spacing accuracy therefore remains to be established.

\subsection{Comparison with prior work}
Domain differences limit direct comparison with existing crack segmentation
methods. Most published benchmarks use concrete, pavement, or masonry rather
than borehole core images. DeepCrack reports F1 scores of 0.86--0.87 on curated
pavement datasets \citep{liu2019deepcrack}. CrackSeg9k shows that cross-domain
generalization typically reduces segmentation quality by 10--20 percentage
points \citep{kulkarni2022crackseg9k}. Our gated fusion model achieves an F1
score of 0.860 (Table~\ref{tab:segmentation_results_comparison}), which is
comparable to these infrastructure crack benchmarks. The result is notable
because the core images contain irregular illumination, surface labels, and
thin cracks across an extreme aspect ratio. We compare F1 rather than IoU
because published IoU values are averaged over classes, whereas ours is
crack-class only. These measures are not interchangeable. CrackCLIP
\citep{liang2025crackclip}, which was designed for weakly supervised crack
segmentation, achieves a crack-class IoU of 0.30 on our data. This result suggests that
prompt-based vision-language models require adaptation to geological textures.

\subsection{Comparison of Segmentation Architectures}
Table~\ref{tab:segmentation_results_comparison} compares the individual
segmentation models with two learned fusion architectures. Mask R-CNN alone
achieves high recall (0.906) but low precision (0.280), resulting in a
crack-class IoU of 0.272. The three-encoder U-Net described in
Section~\ref{sec:unet3enc}, which combines the RGB image, PiDiNet edge map, and
Mask R-CNN mask using image-level source weights, achieves an IoU of 0.639. The
gated U-Net using PiDiNet and Mask R-CNN inputs achieves the highest observed
result, with an IoU of 0.754, precision of 0.816, recall of 0.908, and F1 of
0.860.

The two fusion architectures differ in their inputs, encoder initialization,
depth, capacity, and fusion mechanism. Their performance difference therefore
does not isolate the causal effect of spatial gating. The comparison shows that
the evaluated gated configuration performs best under the present protocol; a
controlled ablation with matched inputs and capacity would be required to
attribute the difference specifically to the gate.

\subsection{When fusion degrades its input}
\label{sec:fusion_fails}
The gate is not universally beneficial. Replacing Mask R-CNN with CrackCLIP as
the mask source reduces crack-class IoU to 0.215
(Table~\ref{tab:segmentation_results_comparison}). This result is lower than the
0.300 achieved by CrackCLIP alone, so fusion degrades its own input in this
setting.

The attention gate learns a spatially varying representation from the
segmentation and edge inputs; it does not explicitly encode agreement between
them. With Mask R-CNN, the high-recall segmentation mask and the PiDiNet edge
map provide complementary evidence, and the fused model raises precision to
0.816 while retaining a recall of 0.908. With CrackCLIP, recall falls from 0.500
for the standalone model to 0.364 after fusion. This result shows that the
learned fusion configuration is sensitive to the quality and sparsity of its
input masks. It does not, by itself, establish the internal gate values or an
agreement-based failure mechanism; saved gate maps would be needed for that
analysis.

\subsection{Failure modes}
Three recurring patterns limit crack segmentation. First, the model often
misses hairline cracks narrower than three pixels. These cracks are
under-represented in the annotations and can run parallel to surface scratches.
Second, cracks separated by fewer than 5--10 pixels can merge into one
predicted region, collapsing distinct spacing intervals. Third, dark brown or
black lithologies can provide weak PiDiNet responses for low-contrast cracks,
causing false negatives at the gate input. These segmentation errors can
propagate into the number and length of intervals produced by post-processing, but that propagation is not quantified in the present study.
Bedding angle estimation also fails when core is missing or heavily fragmented because
these conditions violate the cylindrical geometry assumption. Future work
should address these limitations through multi-scale annotation,
contrast-adaptive preprocessing, and learned geometric priors.

\subsection{Limitations}
\label{sec:limitations}
Three limitations bound how the reported numbers should be read.

First, \textbf{the reference is the log report, not the rock}. Every reference
label comes from observations entered by a geologist into the logging software.
This limitation applies to defect spacing, bedding angle, weathering, and
lithological color. The reported accuracy therefore measures agreement with a
geologist's recorded judgment. As noted in the Introduction, this judgment has
substantial inter-observer variability \citep{alejano2025rock}. That variability
is part of the ground truth and imposes an upper bound that cannot be quantified
without independently re-logging the same core. Even perfect agreement would
reproduce one geologist's observations rather than independently measure the
rock. The limitation does not invalidate relative model comparisons because all
models use the same reference. However, the 75.4\% agreement for bedding-angle categories should not be
interpreted as independently measured physical accuracy. The segmentation reference also treats label-touching cracks as background, so the reported pixel metrics include this source of annotation noise.

Second, \textbf{the corpus is narrow}. All data are from Australia, cover a small number of rock types, and follow one image-acquisition protocol. Several pipeline components depend on this
protocol. Core image extraction detects a particular yellow ruler, defect
spacing extraction assumes one log report template, and the color references
are calibrated to the observed lithologies. The methods are not inherently
site-specific, but the reported results characterize only this corpus.
Generalization to other operators, log report templates, acquisition protocols, and
rock types remains untested.

Third, \textbf{the learned-model results are single-run estimates}. We do not
report variation across random seeds or confidence intervals. The model rankings
therefore describe the observed runs under the fixed borehole-grouped split;
repeated training would be required to quantify their stability.

\section{Conclusion}
We presented a hybrid framework for borehole core analysis that uses report-derived weak labels for defect-spacing classification and manually annotated masks for fully supervised crack segmentation. Among the evaluated segmentation configurations, the gated U-Net using PiDiNet and Mask R-CNN inputs achieved the highest crack-class IoU and F1. Deterministic post-processing converts the predicted crack geometry into spacing intervals, while separate rule-based procedures produce core-relative bedding-angle and lithological color descriptors. Future work will evaluate more diverse acquisition conditions and validate the outputs
against independent geological measurements.

\section*{Data availability}
The borehole log reports and derived core images analyzed in this study are
proprietary and cannot be released. The source code developed for this work is
also not publicly available.

\bibliographystyle{cas-model2-names}
\bibliography{references}

\end{document}